\documentclass{bmvc2k}
\usepackage[]{siunitx}
\usepackage[utf8]{inputenc}

\usepackage{algorithm}
\usepackage[noend]{algpseudocode}

\title{Instance Segmentation of Fibers \\ 
from Low Resolution CT Scans \\
via 3D Deep Embedding Learning}

\addauthor{Tomasz Konopczyński$^\ast$}{Tomasz.Konopczynski@tooploox.com}{13}
\addauthor{Thorben Kröger}{Kroeger@volumegraphics.com}{2}
\addauthor{Lei Zheng}{Lei.Zheng@medma.uni-heidelberg.de}{1}
\addauthor{Jürgen Hesser}{Juergen.Hesser@medma.uni-heidelberg.de}{1}

\addinstitution{
 Department of Radiation Oncology\\
 University Medical Center Mannheim\\
 Heidelberg University, Germany
}
\addinstitution{
 Volume Graphics\\
 Heidelberg, Germany
}
\addinstitution{
 Tooploox\\
 Wrocław, Poland
}

\runninghead{Konopczyński \protect\etal}{Fiber Instance Segmentation from CT via DL}

\def\etal{\emph{et al}\bmvaOneDot}

\begin{document}

\maketitle

\begin{abstract}
We propose a novel approach for automatic extraction (instance segmentation) of fibers from low resolution 3D X-ray computed tomography scans of short glass fiber reinforced polymers.
We have designed a 3D instance segmentation architecture built upon a deep fully convolutional network for semantic segmentation with an extra output for embedding learning.
We show that the embedding learning is capable of learning a mapping of voxels to an embedded space in which a standard clustering algorithm can be used to distinguish between different instances of an object in a volume.
In addition, we discuss a merging post-processing method which makes it possible to process volumes of any size.
The proposed 3D instance segmentation network together with our merging algorithm is the first known to authors knowledge procedure that produces results good enough, that they can be used for further analysis of low resolution fiber composites CT scans.
\end{abstract}

\section{Introduction}

\begin{figure}
\centering
    \includegraphics[width=1.0\linewidth]{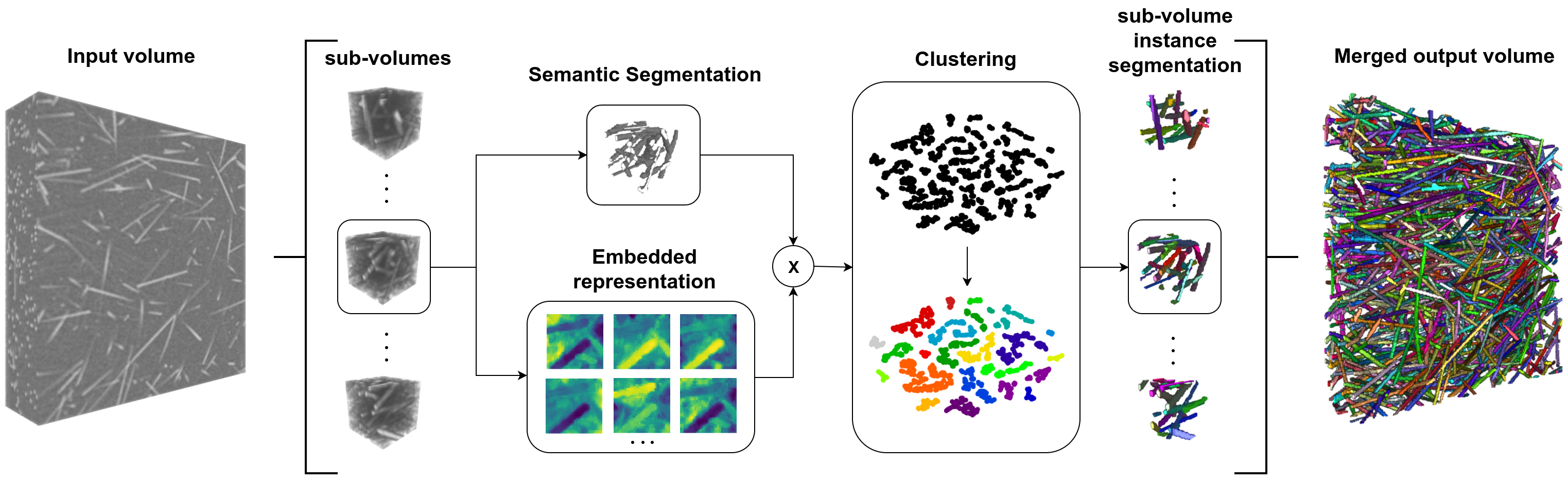}
\caption{
Sketch of the proposed method. The network is processing overlapping sub-volumes of the input volume. For each sub-volume a semantic segmentation mask and an embedding is produced by a deep network. A clustering method is then applied on the segmented regions of the embedding representation producing clusters corresponding to individual fibers. Fibers are then mapped back to the spatial domain.
The overlapping instance sub-volumes are then merged into an output volume.
}
\label{fig:long1}
\end{figure}

Reliable information about fiber characteristics in short-fiber reinforced polymers (SFRP) 
is much needed for process optimization during the product development phase. The influence of fiber characteristics on the mechanical properties of SFRP composites is of particular interest and significance for manufacturers~\cite{fiber1}. The recent development of X-ray computed tomography (CT) for nondestructive quality control enabled the possibility to scan the materials and retrieve the 3D spatial information of SFRPs.
Fiber extraction is the first step towards any further analysis of a SFRP material. 
However, the spatial resolution of a scan is a limiting factor which makes fiber extraction a difficult problem.

Acquiring scans in high resolution is time consuming and costly.
Therefore, in this work we consider only scans acquired by a CT system with low (\SI{3.9}{\micro\meter}) resolution.
The methods currently in use are usually based on hand designed features. Since fibers can be described as long cylindrically shaped objects, the most widely used family of fully-automatic methods is based on Hessian eigenvalues. Using a set of Hessian based filters at a number of scales, a confidence map of fiber occurrence can be produced~\cite{Frangi}. 
To extract individual fibers, a template matching~\cite{Pinter}~\cite{Similar2Pinter} or a watershed splitting and skeletonisation technique~\cite{watershedFiber1}~\cite{watershedFiber2} is then applied. However, the performance of these methods degrades severely if the resolution is too low and fails to produce meaningful results~\cite{KonopReference}. A deep learning method has already shown its superiority over Hessian based techniques to produce more accurate results for semantic segmentation of fibers at low CT resolution~\cite{KonopDeep}.

Deep learning architectures have been successfully applied to semantic segmentation problems for both natural 2D images and 3D CT volumes~\cite{FCN}~\cite{CTLiver}. Similar solutions have been found for the problem of 2D instance segmentation. Faster R-CNN~\cite{FRCNN} and the Mask R-CNN~\cite{MaskRCNN} architectures are examples of region-proposal-based techniques which are the state-of-the-art for common scene-understanding datasets like COCO~\cite{COCO} or ImageNet~\cite{ImageNet}.
However, it is not clear how this approach can be extended to 3D volumetric data with densely packed objects like fibers in SFRP.
This is why for our 3D problem, we have opted for alternative deep learning methods for
instance segmentation.
There are numerous works in which authors try to come up with different ideas for 2D datasets.
An interesting idea that could be extended to 3D
volumes has been proposed by~\cite{WatershedDeep} to reformulate the problem of instance segmentation into
learning a mapping to watershed energy. Then, for the final output, a Watershed transform
is applied to get the instances. Unfortunately, this method is not applicable to our problem,
because fibers are usually too thin to find a border.
Another promising idea proposed by~\cite{TrackingDeep} is to combine convolutional neural networks (CNN) with recurrent neural networks (RNN). The recurrent structures are used to keep track of objects that have already been found, and excludes these regions from further analysis by the algorithm.

In this work we propose a novel deep learning architecture for automatic extraction (instance segmentation) of
fibers from low resolution 3D X-ray computed tomography scans of short glass fiber
reinforced polymers. The sketch of the method is presented in Fig. \ref{fig:long1}.

We explore and discuss the performance of the presented method achieved by training on a low resolution SFRP CT scan and compare it to a standard watershed splitting and skeletonisation technique.
We test the importance of the semantic segmentation branch by replacing it with a ground truth semantic segmentation. 
To the best of our knowledge, this is the first attempt of using deep embedding learning for the task of instance segmentation on a 3D volumetric data. The proposed method is also the first to successfully retrieve single-fiber segmentation from a low resolution SFRP CT scan, while the outcome of the standard methods is producing unacceptable results.
We base our method on an embedding learning approach~\cite{WEIN} ~\cite{KUL}.
The idea is to use a special embedding layer which is placed at the end of a given deep network.
The network is then trained by using a special loss function on the final embedding layer which encourages special structure in the embedding space: pixels belonging to the same class should be close, whereas pixels belonging to different classes should be far apart (in the Euclidean metric of the embedding space). 
The method as a learnable loss function has been first mentioned by~\cite{WEIN}, and was then used with some modifications in the deep learning architecture of~\cite{KUL} and ~\cite{notKUL}. These methods achieved competitive performance on 2D datasets compared with the R-CNN based state-of-the-art. 

In the problem of fiber segmentation, the network will learn a mapping of each voxel and its surrounding from the input to an embedding space in which voxels belonging to one fiber are separated from voxels belonging to another.
Unfortunately, there is one drawback to this method.
Such a network is capable of processing only one small sub-volume of a volume at a time because of memory limitation.
Each time the network processes a sub-volume it assigns an \emph{arbitrary} index to a fiber region.
Because of that, we can not do a simple merge as it is usually done for a semantic segmentation problem, where the output is a probability of being an object of a certain class. For a semantic segmentation mask one can take a simple average over overlapping regions in order to merge sub-volumes into a full volume.
Therefore, in order to produce an instance segmentation for a full CT scan, we propose a  post-processing
algorithm, which merges the overlapping predictions of small blocks into a consistent 
prediction for the entire CT scan during the prediction phase.

\section{Method}

\begin{figure}
\centering
\begin{tabular}{c c c c}
\bmvaHangBox{\includegraphics[width=2.4cm]{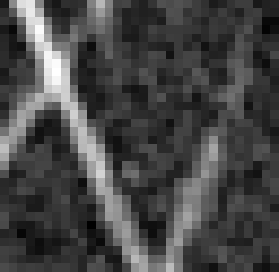}}&
\bmvaHangBox{\includegraphics[width=2.4cm]{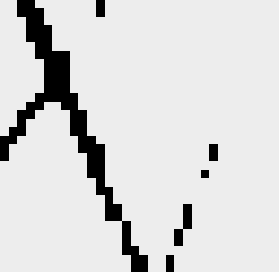}}&
\bmvaHangBox{\includegraphics[width=2.4cm]{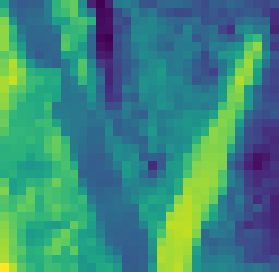}}&
\bmvaHangBox{\includegraphics[width=2.4cm]{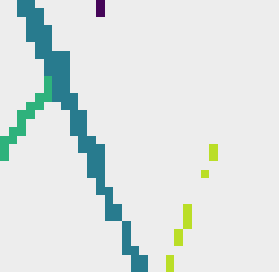}}\\
(a)&(b)&(c)&(d)
\end{tabular}
\caption{Visualization of the network outputs.
(a) Slice of an input volume patch.
(b) Corresponding slice from the semantic segmentation output.
(c) Corresponding slice of one (out of many) feature map of the embedding output.
The network learns to assign different, unique values (colors) to individual fibers.
In  this particular feature map all fibers are well separated.
(d) The masked embedding by a semantic prediction for which a loss for embedding learning is computed.
In the prediction phase it is also the input for the clustering step.
}
\label{fig:branches}
\end{figure}

Similar to the work of~\cite{KUL}, we have extended the Fully Convolutional Network (FCN) architecture~\cite{FCN} designed for semantic segmentation tasks to produce embeddings by using an extra output. 
The extra output could be attached at the very end of the backbone of the semantic segmentation network as in~\cite{KUL2}, but in our setup we decided to use two sub-networks. 
One is responsible for computing the semantic segmentation mask, and the other for computing the embedding of voxels.
The network can be trained separately for embedding and semantic segmentation using corresponding outputs or trained together for both tasks at the same time.
We will refer to the two sub-networks as \textit{semantic segmentation branch} and \textit{embedding learning branch}.
The semantic segmentation branch outputs a confidence map that a given voxel belongs to any fiber or not. 
The embedding learning branch outputs voxels coded in the embedding space.
During the training phase, the architecture is trained only based on outputs from the semantic segmentation branch and the embedding learning branch using specified loss functions. 

\begin{figure}
\centering
    \includegraphics[width=1.0\linewidth]{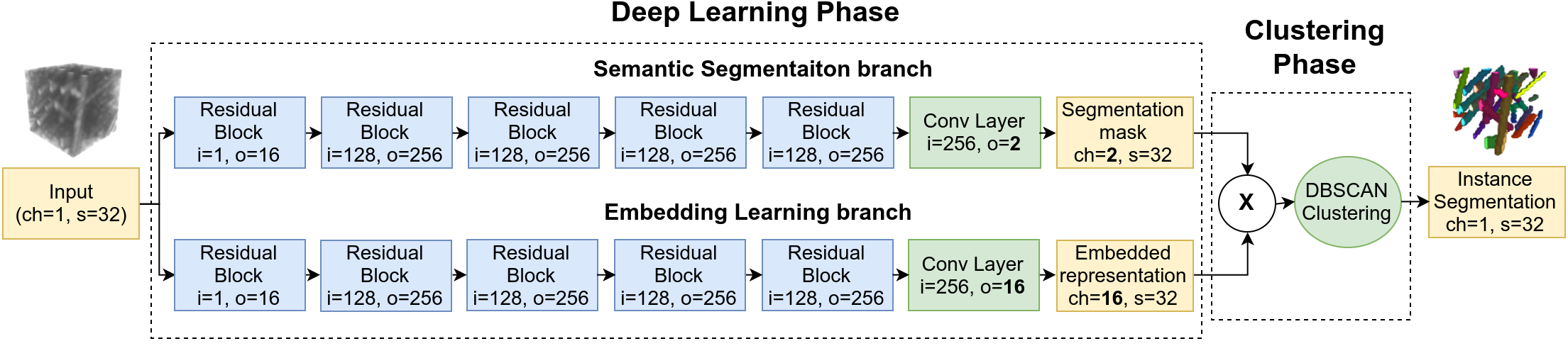}
\caption{
Detailed architecture of our network. \emph{ch} is the number of channels of the sub-volume, \emph{s} is the size of the sub-volume (here $\emph{s}=32$ means a sub-volume is of size $32 \times 32 \times 32$), \emph{i} is the number of input channels, whereas \emph{o} is the number of output channels from a convolutional layer or residual block.
Both residual blocks and convolutional layers use 3D kernels.
The kernel size is set to 3 for residual blocks and 1 for the final convolutional layer. 
Training is performed only on the deep learning phase. During prediction, the output from the embedding branch is masked by the output of the semantic segmentation branch and is processed by the DBSCAN algorithm producing the instance segmentation. The cross sign indicates the masking operation. 
}
\label{fig:arch}
\end{figure}

During the prediction phase a clustering step generates clusters corresponding to individual fibers.
The clustering method is applied to the embedded voxels which have a high confidence of being a fiber based on the output from the semantic segmentation branch. The outputs from the two branches and the region on which the clustering is computed are presented in Fig \ref{fig:branches}.
The clusters are then mapped back to the spatial domain creating a label volume, where each voxel is assigned an integer label
corresponding to the fiber instance it is a part of.
To make it possible to use on volumes of any size, we have proposed a greedy merging algorithm.
The network produces outputs for overlapping sub-volumes of the input volume, which are then merged to a full volume.
The detailed architecture of the network is presented in Fig. \ref{fig:arch}.  In the following sections we will describe the above steps in more detail.
%
\subsection{Semantic Segmentation}
The semantic segmentation branch is a standard FCN for semantic segmentation.
We have used an architecture that has been designed for the task of semantic fiber segmentation~\cite{KonopDeep}.
The output of the branch is penalized by the standard voxel-wise binary cross entropy loss $L_{CE}$, as is common for semantic segmentation tasks. It  is defined as:
\begin{equation} 
L_{CE} =  -  [y \cdot log(\hat{y})  + (1-y)\cdot log(1-\hat{y})]
\end{equation}
where $y$ are the true binary labels, and $\hat{y}$ are the predicted labels. 
During the prediction phase, the output is thresholded at value 0.5 in order to produce binary masks.
An example slice of an output of the branch is shown in Fig \ref{fig:branches} (b).

\subsection{Embedding Learning Loss}
\begin{figure}
\centering
    \includegraphics[width=0.95\linewidth]{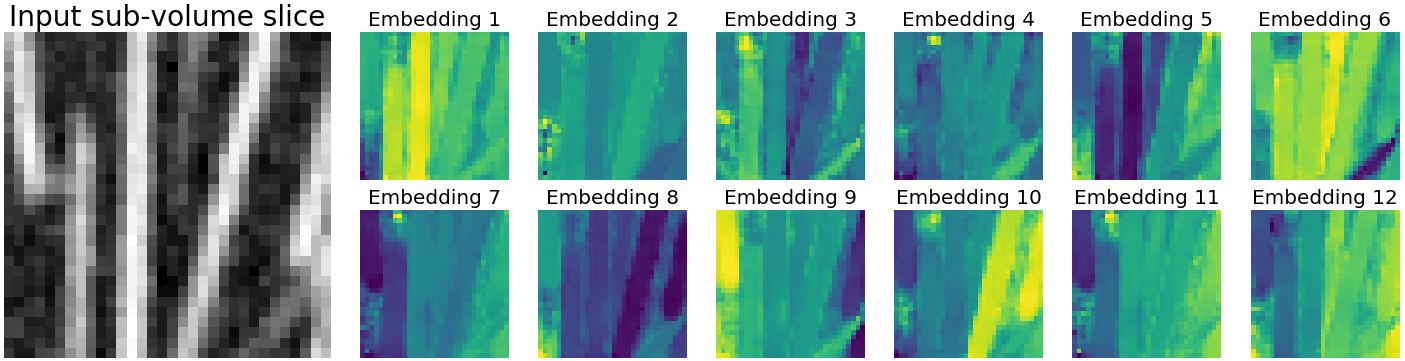}
\caption{
The input volume is represented by a number of embedded volumes at the embedding branch output. Here, slices of the first 12 embedding volumes corresponding to the input sub-volume slice are visualized. Note, that a good embedding will assign different set of colors in each embedding volume so that the clustering in the embedding space will be easy.
}
\label{fig:colorful}
\end{figure}
The output of the embedding branch is a representation of the sub-volume in an embedding space.
The architecture of the branch is identical to the semantic segmentation branch.
The only difference is the number of output channels in the final convolutional layer and the loss function.
In the semantic segmentation task, the output is producing a volume with two channels, where one is reasoning on the foreground and the other on the background.
In the embedding learning, the output has as many channels as the dimensionality of the embedding space (a hyperparameter in the algorithm). An example visualizing feature maps of the embeddings is shown in Fig. \ref{fig:colorful}. 

The loss function penalizes voxels of different instances that are too close to each other in the embedding space and encourages voxels of the same instance to be close.
As a result, the network maps the voxels into the embedding space, such that voxels that belong to the same fiber should be placed next to each other and form easily separable clusters.

We find that the loss function introduced by~\cite{KUL} inspired by work of~\cite{WEIN} and extended to 3D by us works best for our problem. Even though we have extended the problem to 3D, and have used data that contains a high number of objects compared to common scene-understanding problems, the method does not seem to be affected by that.
The loss consists of three terms: $L_v$ keeps voxels belonging to the same object close to each other, $L_d$ which forces a minimal distance between clusters of different objects, and $L_r$ which regularizes the cluster centers to be close to the origin. The terms are defined as:
\begin{align} 
L_{v} &=  \frac{1}{C} \sum_{c=1}^{C} \frac{1}{N_c} \sum_{i=1}^{N_c} [|| \mu_c - x_i  || - \delta_v ]_+ ^2   \\
L_{d} &=  \frac{1}{C(C-1)} \sum_{c_A=1}^{C}  \sum_{c_B=1, c_A \neq c_B}^{C} [\delta_d - || \mu_{c_A} - \mu_{c_B}  ||]_+ ^2  \\
L_{r} &=  \frac{1}{C} \sum_{c=1}^{C}|| \mu_c || 
\end{align}
where $C$ is the number of objects in the ground truth patch (clusters),
$N_c$ is the number of voxels that corresponds to the object $c$,
$x_i$ is the embedding in the final embedding layer,
$\mu_c$ is the mean of the embedding of object $c$, 
$||\cdot||$ is the $L_2$ norm, and
$[x]_+ = max(0,x)$.
The parameters $\delta_v$ and $\delta_d$ are used to control the desired positions of the clusters. The final loss for the embedding learning $L_{embd}$ is a sum of the previous components.
\begin{equation} 
L_{embd} =  \alpha L_{v} + \beta L_{d} + \gamma L_{r}
\end{equation}
where $\alpha, \beta$ and $\gamma$ control the strength of the corresponding term.
An example slice of an output of the branch is visualized in Fig \ref{fig:branches} (c).
Note, that the loss is computed only based on the voxels that belong to the foreground fibers. It is the task of the semantic segmentation branch to find the correct position of the fibers.

\subsection{Clustering}
%
%
%
%
\begin{figure}
\centering
\begin{tabular}{c c c c}
\bmvaHangBox{\includegraphics[width=2.84cm]{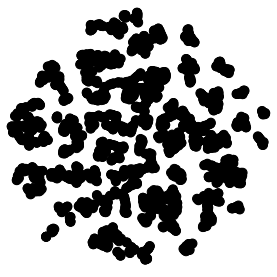}}&
\bmvaHangBox{\includegraphics[width=2.84cm]{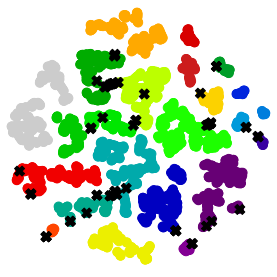}}&
\bmvaHangBox{\includegraphics[width=2.84cm]{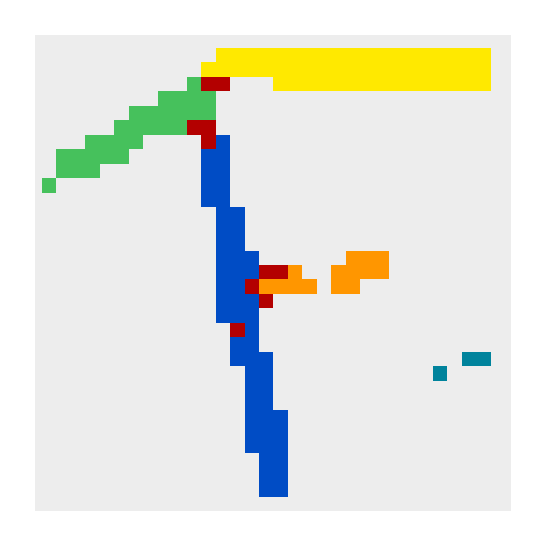}}&
\bmvaHangBox{\includegraphics[width=2.84cm]{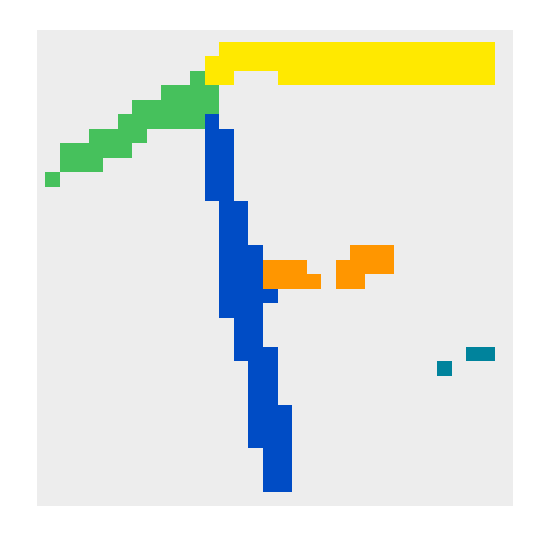}}\\
(a)&(b)&(c)&(d)
\end{tabular}
\caption{1. Visualization of the clustering steps of the method.
(a) Masked embeddings form clusters in a multi-dimensional embedding space (visualized by t-SNE).
(b) DBSCAN clusters the embedding representation and assign a different index (color) to each fiber (cluster) with black crosses for outliers.
(c) Clusters are then mapped back to the spatial domain. Here a corresponding example slice of the mapping with red pixels for outliers.
(d) The watershed algorithm is then applied as a post-processing step to fill the outliers.
}
%
\label{fig:DBSCAN}
\end{figure}


As discussed in the previous section, the semantic segmentation output creates a confidence map that a given voxel belongs to any fiber or not.
A clustering is then applied to the embedded voxels with a high confidence of being fibers.
An example input slice of one of the feature maps of the embedding is shown in Fig \ref{fig:branches} (d).
In this work, we found DBSCAN~\cite{DBSCAN} to work best on the SFRP dataset.
In contrast to Mean Shift used in~\cite{KUL}, DBSCAN does not make assumptions about the shape of the clusters.
We apply clustering only in the prediction phase because the instance segmentation loss function does not require the instance segmentation map.
Note, that DBSCAN does not necessarily assign a label to all voxels. 
%
Voxels that were not assigned to any label are assigned as outliers.
The clusters are then mapped back to the spatial domain creating an instance segmentation map.
Outliers are extrapolated based on their neighborhood in the spatial domain by use of the watershed algorithm, using the clustering labels as seeds.
An example visualization of the described steps with help of the t-SNE~\cite{tSNE} is shown in Fig \ref{fig:DBSCAN}.

\subsection{Merging}
Finally, the inference is produced on small overlapping sub-volumes of the entire volume.
Each sub-volume contains different label IDs for fibers, making it not clear which fiber is which.
To overcome this problem we have designed a merging algorithm, which joins label IDs among the sub-volumes based on a spatial distance of fibers in the overlapped regions.
The algorithm is applied at each sub-volume and processes recursively one fiber at a time, looking at neighboring sub-volumes with overlapping regions with objects being close to the fiber of interest. The merging procedure is described in more details in algorithm 1.

\begin{algorithm}
\caption{Merging algorithm}\label{alg:merge}
\begin{algorithmic}[1]
\Procedure{Merge}{$f,p$} \Comment{for a fiber $f$ in a sub-volume $p$}
\State $N\gets$ neighbour sub-volume of $p$ \Comment{$N$ is a set of sub-volumes neighbouring with $p$}
\For{sub-volume $n$ in $N$} 
    \State $G\gets$ fibers in $n$ \Comment{$G$ is a set of fibers in a patch $n$}
    \For{fiber $g$ in $G$} 
         \State $d\gets$ $D(f,g)$ \Comment{Spatial distance between $f$ and $g$}
        \If{$d> \alpha$} 
            \State $g_{id}\gets f_{id}$
            \State \Call{Merge}{$g, n$}
        \EndIf
    \EndFor
\EndFor

\EndProcedure
\end{algorithmic}
\end{algorithm}

\section{Experiments}

\subsection{Data}

We have evaluated the proposed setup on two hand-annotated regions of low resolution CT scans of SFRP composites acquired by a Nikon MCT225 X-ray CT system from~\cite{KonopReference}.
Scans exhibit typical artifacts and have low, but isotropic resolution.
The parts from which the scans were acquired were manufactured by micro injection molding using PBT-10\% GF, a commercial polybutylene terephthalate PBT (BASF, Ultradur B4300 G2) reinforced with short glass fibers (10\% in weight).
The volumes have been hand annotated with center lines and processed by a watershed algorithm to create the instance segmentation ground truth.
Both volumes are cubes of dimension 62 $\times$ 260 $\times$ 260 with approx. 6,500 fibers each.
Fibers have a diameter of 10-14 \SI{}{\micro\meter} (2-3 voxels)  and are approx. 1.1 mm long.
One scan is used for training, while the other is only used for testing.

\begin{figure}
\begin{center}
\begin{tabular}{c | c | c | c}
Raw volume & Ground truth & CC & Our method \\
\bmvaHangBox{\includegraphics[width=0.21\linewidth]{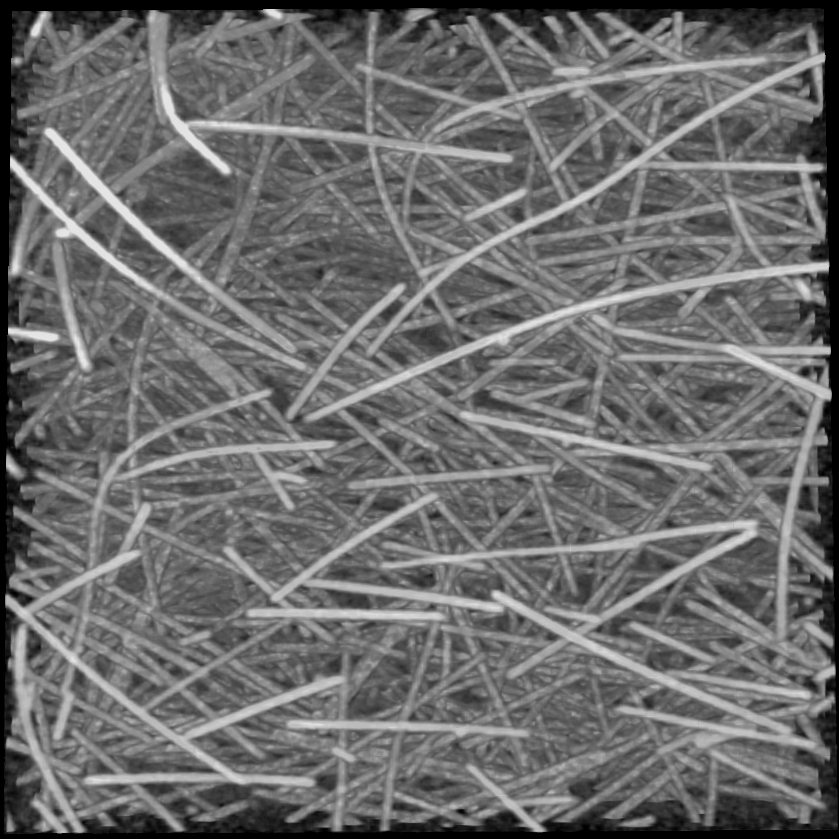}}&
\bmvaHangBox{\includegraphics[width=0.21\linewidth]{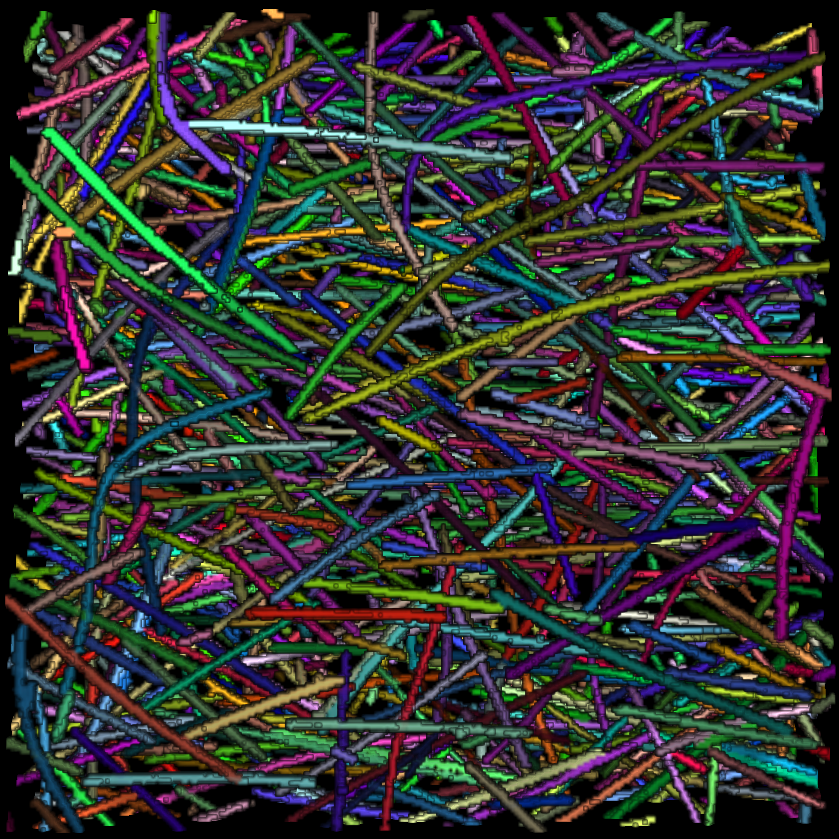}}&
\bmvaHangBox{\includegraphics[width=0.21\linewidth]{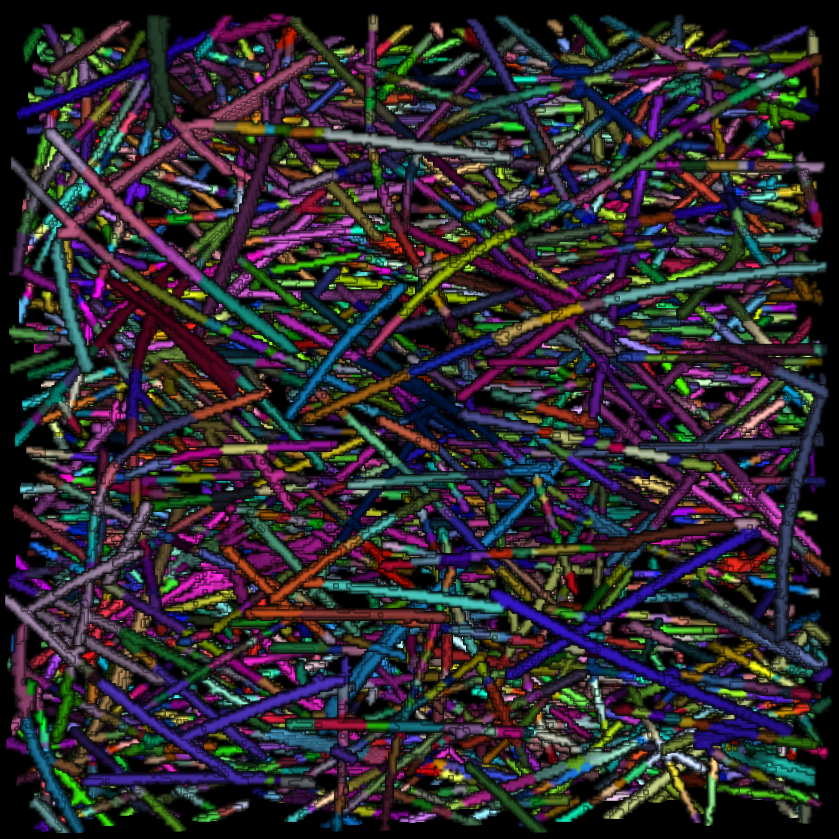}}&
\bmvaHangBox{\includegraphics[width=0.21\linewidth]{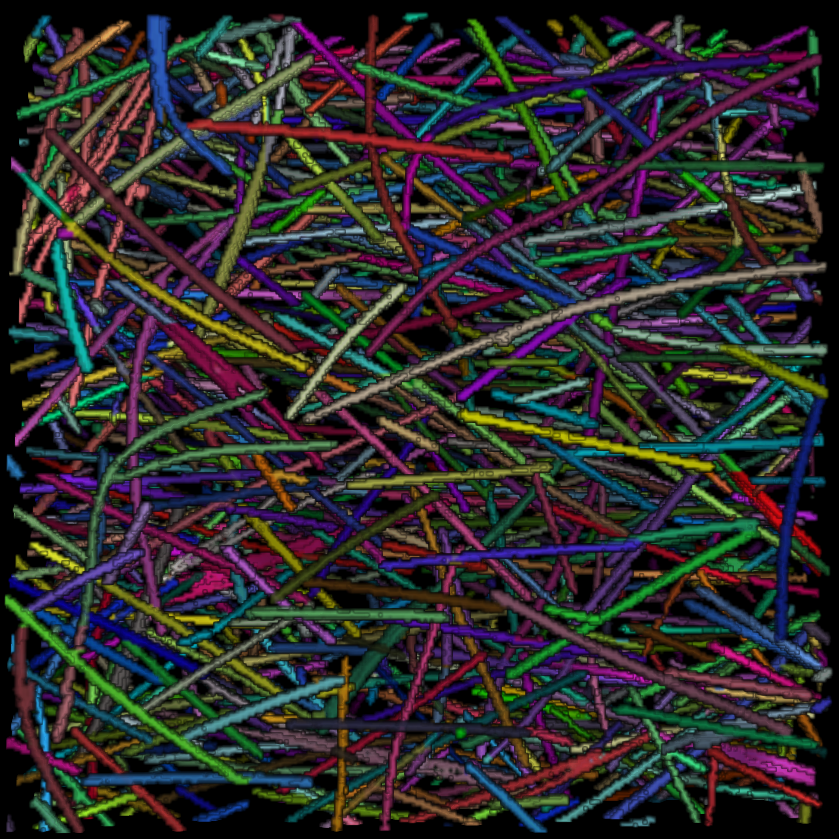}}\\
\bmvaHangBox{\includegraphics[width=0.21\linewidth]{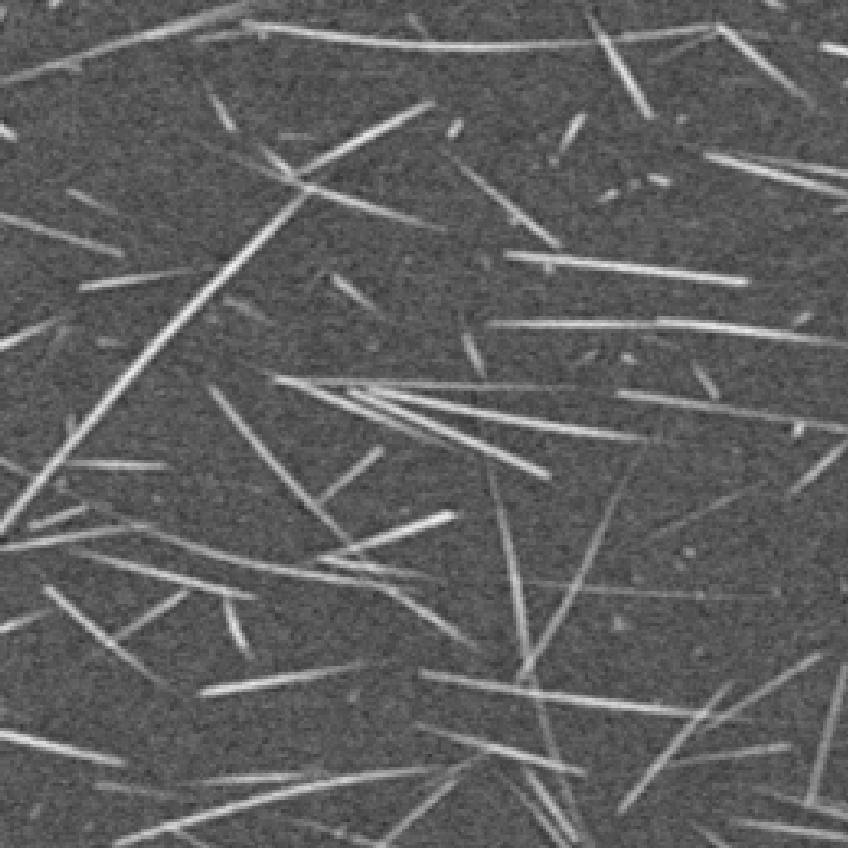}}&
\bmvaHangBox{\includegraphics[width=0.21\linewidth]{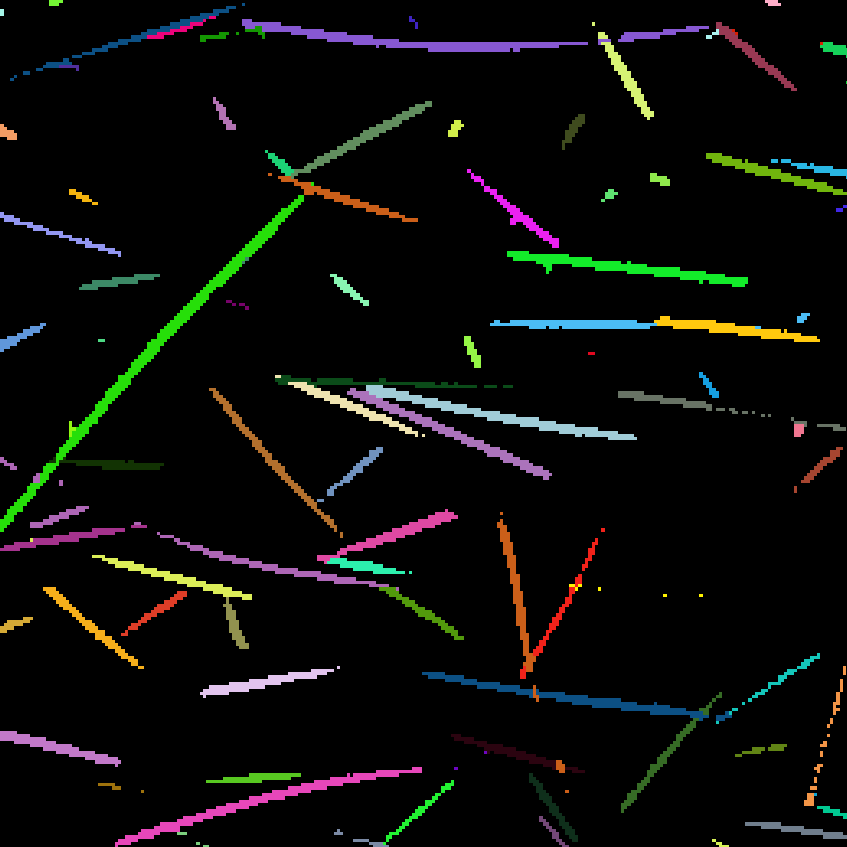}}&
\bmvaHangBox{\includegraphics[width=0.21\linewidth]{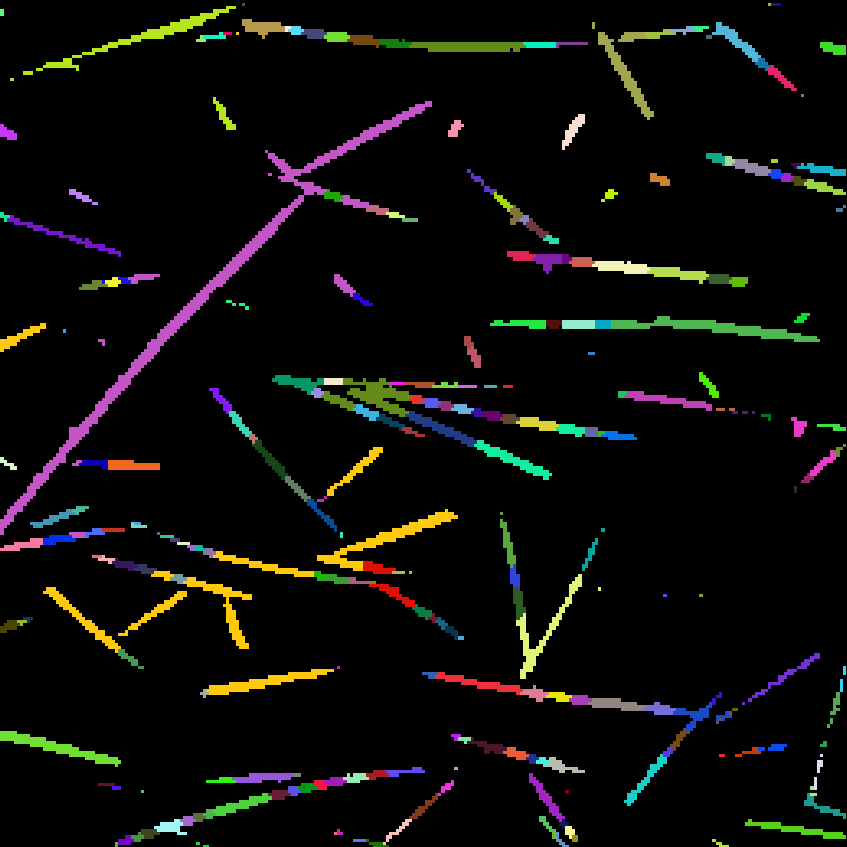}}&
\bmvaHangBox{\includegraphics[width=0.21\linewidth]{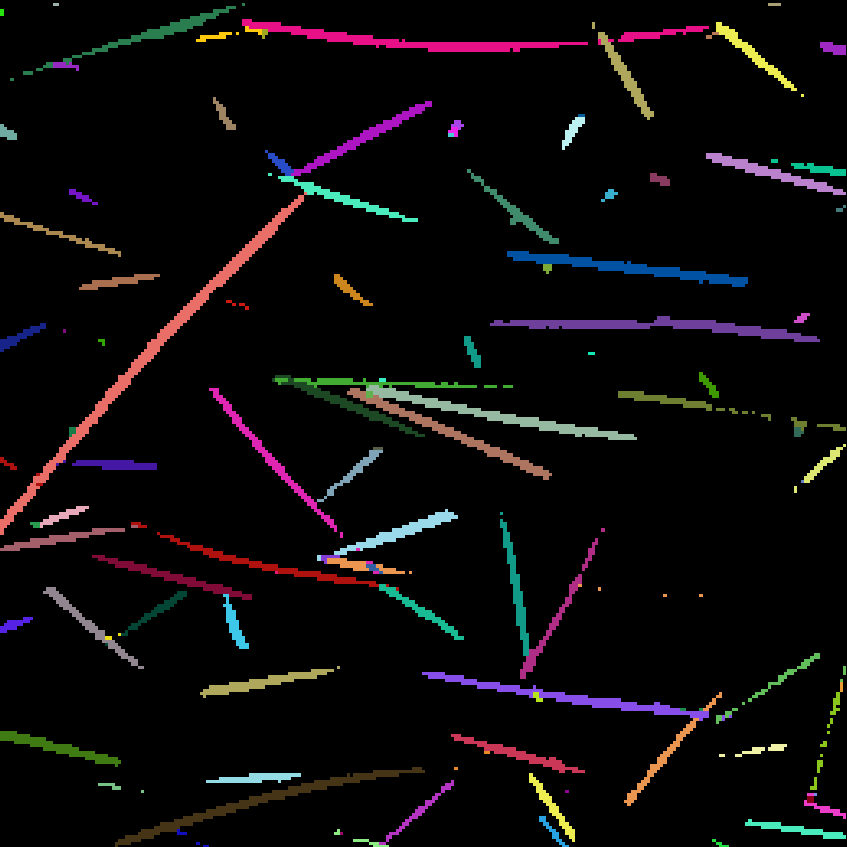}}\\
\end{tabular}
\end{center}
\caption{Visualization  of  the  testing  volume and corresponding results. First row shows a 3D rendering of a volume. Second row shows one example slice of the same volume.
First column is the input test volume (with a certain threshold to remove the epoxy background in the 3D rendering). Second column is the corresponding ground truth. Third column is the output of a standard connected component (CC) analysis. Fourth column is the output of our method. Fibers are colored semi-randomly based on the fiber ID.
}
\label{fig:final}
\end{figure}


%
%

%
%

\subsection{Training details}
The volumes have been normalized to have unit variance and zero mean.
Additionally, most of the air voxels surrounding the specimen have been removed by a simple thresholding method.
We have trained and evaluated the network on sub-volumes of $32 \times 32 \times 32$ from the training volume.
The sub-volumes are randomly flipped and rotated (by 90, 180 or 270 degrees) during the training phase.
As mentioned in the introduction, and shown in Figure 3, for backbones of both the semantic segmentation and embedding learning branch we have used the architecture proposed in~\cite{KonopDeep} designed for semantic fiber segmentation.
It is a 3D FCN with standard residual units~\cite{RES} and batch normalization~\cite{BATCH} but with no max-pooling to keep the resolution of the already very thin fibers.

The embedding learning is not stable, when trained from noise.
Therefore, first we have trained the semantic segmentation branch for 20,000 iterations and saved the weights.
Then, we have used the weights as an initialization for the embedding learning branch and trained it for another 20,000 iterations. 
The loss used for training the embedding learning uses the semantic ground truth masks.

It would also have been an option to share the embeddings and weights for both tasks.
Such setup is reported to slightly increase the performance of both semantic and instance segmentation~\cite{KUL2}. However, in our setup, we have found the above two-stage training to work better.
We use 16 feature embedding maps and set $\alpha$ and $\beta$ to 1 and $\gamma$ to 0.001.
Optimization has been done by using the Adam optimizer~\cite{Adam} with an initial learning rate set to 0.001.
During the prediction phase, the algorithm processes overlapping $32 \times 32 \times 32$ sub-volumes of the test volume with an overlap of $16$ in each direction.
The post-processing merging algorithm merges the overlapping sub-volumes and produces the final instance segmentation volume.

For a metric we have use the Adjusted Rand Index~\cite{ARI} to measure the performance of instance segmentation. We find it more informative in the context of SFRP data over the mAP.
Defining the ground truth labels as clusters $C = \{ C_1, ..., C_k \}$ and the corresponding predicted labels as clusters $\hat{C} = \{ \hat{C}_1, ..., \hat{C}_l \}$, the Adjusted Rand Index $R_a$ is:
\begin{equation} 
R_a(C, \hat{C}) = \frac{\sum_{i=1}^{k} \sum_{j=1}^{l} \binom{m_{ij}}{2} -t_3}{\frac{1}{2} (t_1 + t_2) -t_3}
\end{equation}

where $m_{ij} = |C_i \cap \hat{C}_j|$, 
$t_1 = \sum_{i=1}^{k} \binom{|C_i|}{2}$, 
$t_2 = \sum_{j=1}^{l} \binom{|\hat{C}_j|}{2} $, 
$t_3 = \frac{2 t_1 t_2}{n(n-1)}$, and $n$ is the number of voxels in the volume.
The Rand Index varies from 0 to 1, where 1 means a perfect match between the algorithm output and the ground truth mask.

\subsection{Results}

\begin{table}
\begin{center}
\begin{tabular}{|c|c|c|}
\hline
 \textbf{Setup} & \textbf{Mean ARI} & \textbf{Merged ARI}  \\
\hline
Embedding Learning & 0.9048 & 0.6529  \\
\hline
Embedding Learning + true semantic & \textbf{0.9129} & \textbf{0.7817} \\
\hline
Connected Components & 0.3537 & 0.2112   \\
\hline
Connected Components + true semantic & 0.3614 & 0.2534 \\

\hline
\end{tabular}
\end{center}
\caption{
Comparison of our method with traditional connected components with and without provided ground truth semantic segmentation mask. Mean ARI are mean values over overlapping sub-volumes of the validation volume, while Merged ARI is the score computed over the entire volume after the post-processing merging step. The Dice score of the semantic segmentation mask from the semantic segmentation branch is 0.9784. 
}
\label{tab:table}
\end{table}

%
We have compared our method to a standard skeletonization followed by connected component analysis and the Watershed method~\cite{watershedFiber2}.
In the method, a binary erosion is first applied on the semantic mask, which serves as seeds after connected component analysis for a watershed segmentation algorithm. See Fig. \ref{fig:final} for a visual comparison.
We have also evaluated the importance of a good semantic segmentation mask.
We provide results for both our method and connected components given the semantic segmentation computed by the semantic segmentation branch as well as using the ground truth semantic segmentation.

Therefore we compare four different setups. Our \textit{Embedding Learning} method using the final instance segmentation produced given the semantic segmentation mask from the semantic segmentation branch.
 \textit{Embedding Learning + true semantic} which is our method but using ground truth semantic segmentation mask instead of the one produced by the methods branch (which is not ideal).
 \textit{Connected Components} and  \textit{Connected Components + true semantic}  is the connected component method used either on the output of the semantic segmentation branch or the ground truth semantic mask.

We provide two results in Table \ref{tab:table} for each setup. In the first column, the mean ARI is the mean ARI of all the sub-volumes in the test volume without the merging step.
In the second column one can see the score computed over the entire volume after the post-processing merging step which we call a merged ARI.
We report the ARI score only for the voxels that belong to the ground truth instance segmentation mask.
Including the background voxels would artificially increase the score. 

While the standard method clearly fails even when using the true semantic segmentation mask, the proposed method produces meaningful results in all cases.
When reasoning on small overlapping patches the proposed method achieves 0.9048 average ARI score.
The merging algorithm has trouble with ambiguity of two neighboring outputs and favors merging over splitting.
This results in merging two fibers into one, when they are too close to each other.
After the merging post-processing step the ARI score decreases to 0.6529.

\section{Conclusions}
In this work, we proposed a deep 3D fully convolutional architecture together with a set of post-processing steps for a problem of single fiber segmentation from CT scans of SFRP.
We extend a less common approach of embedding learning for the task of 3D instance segmentation.
We explain in detail the steps of the method together with a post-processing and a merging procedure.
We show that we are better than the traditional skeletonization - watershed method.
We expect our findings to be applicable to a wide variety of volumetric data and not only to fiber composites.


\bibliography{egbib}
\end{document}